%% file: main_arxiv.tex
\documentclass[a4paper, 
10pt
]{article}

\usepackage[a4paper,left=3cm,right=3cm,top=2.5cm,bottom=2.5cm]{geometry}
\usepackage{amsmath}
\usepackage{multirow}
\usepackage[table]{xcolor}
\usepackage{amsopn}
\DeclareMathOperator{\diag}{diag}

\usepackage{subcaption}
\usepackage{pgfplots}
\usepackage{pgfplotstable}
\usepackage{mathtools}
\usepackage{multicol}
\usepackage{comment}
\usepackage{booktabs}
\usepackage{amssymb}
\usepackage{url}
\usepackage{bm}
\usepackage[a4paper,left=3cm,right=3cm,top=2.5cm,bottom=2.5cm]{geometry}
\usepackage{amsmath}
\usepackage{multirow}
\usepackage{float}
\usepackage{tabularx}
\usepackage{enumerate}
\usepackage{array}
\usepackage{xspace}
\usepackage{tikz}
\usepackage[export]{adjustbox}
\usepackage{tikzsymbols}
\usepackage{authblk}

\numberwithin{equation}{section}
\usepgfplotslibrary{groupplots}
\usepgfplotslibrary{fillbetween}
\pgfplotsset{compat=1.5}
\usetikzlibrary{calc,trees,positioning,arrows,chains,shapes.geometric,%
    decorations.pathreplacing,decorations.pathmorphing,shapes,%
    matrix,shapes.symbols, decorations.markings, patterns,fit,quotes,%
    arrows.meta, spy}

\usepackage{xcolor}
\definecolor{Gray}{gray}{0.9}

\usepackage{hyperref}
\usepackage{siunitx}
\usepackage{cleveref}
\usepackage{algorithm}
\usepackage{algpseudocode}
\usepackage[bottom]{footmisc}

\graphicspath{{figures/}}

\newcommand{\mupar}{\ensuremath{\boldsymbol{\mu}}}

\newcommand{\upar}{\ensuremath{\mathbf{u}}}
\newcommand{\R}{\ensuremath{\mathbb{R}}}

\newcommand{\x}{\ensuremath{\mathbf{x}}}
\newcommand{\h}{\ensuremath{\mathbf{h}}}
\newcommand{\z}{\ensuremath{\mathbf{z}}}
\newcommand{\y}{\ensuremath{\mathbf{y}}}
\newcommand{\cpar}{\ensuremath{\mathbf{c}}}

\newcommand{\fullnn}{\ensuremath{\mathcal{ANN}}}

\DeclareMathOperator*{\argmin}{arg\,min}

\newcommand{\V}{\ensuremath{\mathbf{V}}}

\newcommand{\Cpar}{\ensuremath{\mathbf{C}}}

\newcommand{\Lambdapar}{\ensuremath{\boldsymbol{\Lambda}}}
\newcommand{\phipar}{\ensuremath{\boldsymbol{\phi}}}
\newcommand{\Phipar}{\ensuremath{\boldsymbol{\Phi}}}
\newcommand{\alphapar}{\ensuremath{\boldsymbol{\alpha}}}

\begin{document}

\title{A Dimensionality Reduction Approach for Convolutional Neural Networks}

\author[]{Laura~Meneghetti\footnote{laura.meneghetti@sissa.it}}
\author[]{Nicola~Demo\footnote{nicola.demo@sissa.it}}
\author[]{Gianluigi~Rozza\footnote{gianluigi.rozza@sissa.it}}

\affil{Mathematics Area, mathLab, SISSA, via Bonomea 265, I-34136
  Trieste, Italy}

\maketitle

\begin{abstract}
  The focus of this paper is the application of classical model order reduction techniques, such as Active Subspaces and Proper Orthogonal Decomposition, to Deep Neural Networks. We propose a generic methodology to reduce the number of layers of a pre-trained network by combining the aforementioned techniques for dimensionality reduction with input-output mappings, such as Polynomial Chaos Expansion and Feedforward Neural Networks. The necessity of compressing the architecture of an existing Convolutional Neural Network is motivated by its application in embedded systems with specific storage constraints.  Our experiment shows that the reduced nets obtained can achieve a level of accuracy similar to the original Convolutional Neural Network under examination, while saving in memory allocation.
\end{abstract}


\section{Introduction and motivations}
\label{sec:intro}
\input{./sections/intro.tex}

\input{./sections/methods.tex}

\section{The Reduced Artificial Neural Networks}
\label{sec:ASNN}
\input{./sections/ASNN.tex}

\section{Numerical results}
\label{sec:results}
\input{./sections/results.tex}

\section{Conclusions and perspectives}
\label{sec:conclusion}
\input{./sections/conclusion.tex}

\section*{Acknowledgements}
This work was partially supported by an industrial Ph.D. grant sponsored by Electrolux Professional, and was partially funded by European Union Funding for
Research and Innovation --- Horizon 2020 Program --- in the framework
of European Research Council Executive Agency: H2020 ERC CoG 2015
AROMA-CFD project 681447 ``Advanced Reduced Order Methods with
Applications in Computational Fluid Dynamics'' P.I. Professor Gianluigi Rozza.

\bibliographystyle{abbrv}
\bibliography{biblio}

\end{document}

%% file: sections/intro.tex
Neural network is a widespread machine learning technique, increasingly employed in many different fields such as computer vision~\cite{krizhevsky2012imagenet}, natural language processing~\cite{young2018recent}, robotics~\cite{noda2014multimodal}, and speech recognition~\cite{graves2006connectionist}. The accuracy of such models is strictly related to the number of layers, neurons, and inputs \cite{Goodfellow, khan2020survey, trenn2008}, therefore, to tackle even more complex problems, architectures are forced to go deep. If on one hand we have an increasing precision, on the other hand the high number of degrees of freedom results in a longer optimization step and, on a practical side, a bigger architecture to manage. The dimension of the network rarely is considered a bottleneck of this methodology, but the diffusion of neural networks in many engineering fields led to its employment also in embedded systems~\cite{roth2020resource}, that typically show a limited hardware. In these contexts the size of the architecture can be an additional constraint, requiring for a reduction in the number of degrees of freedom of the network.

Finding the intrinsic dimension of neural networks is a very challenging task and, to the best of authors knowledge, not supported by rigorous theoretical proofs. Among the different proposed methods, we mention network pruning and sharing~\cite{han2015deep, liu2018rethinking, liu2018dynamic}, low-rank matrix and tensor factorization~\cite{sainath2013low, zhang2015accelerating, novikov2015tensorizing}, parameter quantization~\cite{courbariaux2016binarized, deng2018gxnor}, and knowledge distillation~\cite{hinton, romero2014fitnets}. In this contribution we propose an extension of the idea explored in~\cite{cui2020active}, where the Active Subspace (AS) property and Polynomial Chaos Expansion (PCE) are exploited to provide a reduced and more robust version of the original network. While such contribution analyzed the AS capability for reducing a deep architecture, we aim here to provide a generic framework for neural network reduction, investigating other mathematical tools rather than AS and PCE.
Mimicking the procedure presented in~\cite{cui2020active}, the original architecture is initially splitted in two cascading parts, the \emph{pre-} and \emph{post-model}: we assume that the second one brings a negligible contribution to the final outcome, giving us the possibility to approximate such part of the model without introducing a larger error. A response surface (or in more general terms, an input-output mapping) is indeed built to fit the data, replacing the last layers of the network. This response surface may belong to an high-dimensional space, since the input dimension is equal to the dimension of the output features of the pre-model. It implies that, to keep the reduction computationally affordable, we also need a dimensionality reduction of the pre-model outputs (which, we remark, are also the input parameters of the response surface). Combining all these ingredients, we obtain a reduced version of the network, containing just few of the initial layers, but with a precision comparable to the full model. We specify that the numerical experiments we are going to present involve only Convolutional Neural Networks (CNNs), but the generality of the methodology allows in principle also for applicability to other models.

We explore in this contribution different tools for the dimensional reduction and for the response surface: in addition to AS and PCE, already tested in the aforementioned reference, we employ Proper Orthogonal Decomposition (POD) and Feedforward Neural Network (FNN). The first is a well established technique for model order reduction \cite{RozzaHessStabileTezzeleBallarin2020, SalmoiraghiBallarinCorsiMolaTezzeleRozza2016, RozzaMalikDemoTezzeleGirfoglioStabileMola2018} that, similarly to AS, compress the data by projecting it onto a low-dimensional space. FNN is instead applied to construct the surface response, as alternative to PCE. The advantage of FNN over PCE is twofold: {\it i},) the simplified input-output mapping (thanks to the low-dimensional space) allows to use a FNN with few layers and neurons, reducing further the already minimal space demand of the PCE method; {\it ii},) on the programming side, the possibility to approximate part of the neural network with another network makes the software integration easier, especially when the hosting system is embedded.

The article is organized as follows. \Cref{sec:methods} provides an algorithmic overview of all the numerical methods involved in the reduction framework (AS in \cref{sec:AS}, POD in \cref{sec:POD}, PCE in \cref{sec:PCE}, and FNN in \cref{sec:ANN}), while in \cref{sec:ASNN} we cover in details the framework to reduce the neural networks.
In \cref{sec:results} we present the results obtained by reducing with the proposed methodology a benchmark CNN designed for image recognition. We repeat such analysis using two different datasets during the initial learning step, investigating the dependency of the results on the original problem. Finally in \cref{sec:conclusion} we summarize the entire procedure and propose some future perspectives to enhance the framework.

%% file: sections/methods.tex
\section{Numerical tools}
\label{sec:methods}
We introduce in this section all the techniques employed for the reduction of the network, in order to make easier to understand the framework in Section~\ref{sec:ASNN}.

\subsection{Dimensionality reduction techniques}
The subsection is devoted to an algorithmic overview of the reduction methods tested within this contribution, the Active Subspace (AS) property and the Proper Orthogonal Decomposition (POD). Widely employed in the reduced order modeling community, such techniques are used to reduce the dimensionality of the intermediate convolutive features, but we postpone to the next section the details. We just specify that, even if in this work we focus on AS and POD, the framework is generic, allowing in principle to replace these two with other methods for reducing the dimensionality.

\subsubsection{Active Subspaces}
\label{sec:AS}
\input{./sections/AS.tex}

\subsubsection{Proper Orthogonal Decomposition}
\label{sec:POD}
\input{./sections/POD.tex}

\subsection{Input--output mapping}
Once the outputs of the intermediate layer are dimensionally reduced, we need to correlate the latter to the final output of the original network, e.g. the belonging classes in an image identification problem. An input--output mapping is then built starting from the input dataset.
The next subsections are dedicated to the algorithmic overview of the two methods explored for approximating the mapping: the Polynomial Chaos Expansion (PCE) \cite{xiu2002wiener} and the Feed--forward Neural Network (FNN) \cite{fine2006feedforward}.
\subsubsection{Polynomial Chaos Expansion}
\label{sec:PCE}
\input{./sections/PCE.tex}

\subsubsection{Feedforward Neural Network}
\label{sec:ANN}
\input{./sections/ANN.tex}

%% file: sections/AS.tex
Active Subspaces (AS) \cite{constantine2015active, Constantine_2014} method is a reduction tool used to identify important directions in the parameter
space by exploiting the gradients of the function of interest. Such information
allows to apply a rotational transformation to the domain in order to obtain an
approximation of the original function in lower dimension.
Let $\mupar = [\mu^1 \dots \mu^n]^T \in \R^n$
represent a $n$-dimensional variable with a probability density function
$\rho(\mupar)$, and let $g$ be the function of interest, $g(\mupar): \R^n \to \R$. We assume here $g$ is scalar and continuous (for the vector-valued extension see~\cite{romor2020kernelbased, zahm2020gradient}). 
Starting from this, an uncentered covariance matrix $\Cpar$ of the gradient of $g$ can be constructed by
considering the average of the outer product of the gradient with itself:
\begin{equation}
    \Cpar =\mathbb{E}[\nabla g(\mupar)\nabla g(\mupar)^T] = \int (\nabla _{\mupar} g)(\nabla _{\mupar} g)^T \rho \textrm{d}\mupar, 
\end{equation}
where the symbol $\mathbb{E}[\cdot]$ denotes the expected value, and $\nabla_{\mupar} g
\equiv \nabla g(\mupar)$. We assume the gradients are computed during the simulation,
otherwise  if not provided they can be approximated with different techniques
such as local linear models, global models, finite difference, or Gaussian
process \cite{abel, williams2006gaussian}, for example. Since $\Cpar$ is symmetric it admits the following eigenvalue
decomposition:
\begin{equation}\label{eq:eigen}
\Cpar = \V \Lambdapar \V^T, \qquad \quad \Lambdapar = \diag(\lambda_1, \dots,\lambda_n), ~~ \lambda_1\ge \cdots \ge \lambda_n\ge 0,
\end{equation}
where $\V$ is the $n \times n$ orthogonal matrix whose columns $\{\mathbf{v}_1, \dots,
\mathbf{v}_n \}$ are the normalized eigenvectors of $\Cpar$, whereas $\Lambdapar$ is a
diagonal matrix containing the corresponding non-negative eigenvalues
$\lambda_i$, for $i=1,\dots, n$, arranged in descending order.\\
We can decompose these two matrices as:
\begin{equation}
\Lambdapar =  
\begin{bmatrix}
\Lambdapar_1 & \\
&\Lambdapar_2
\end{bmatrix},\qquad \V = [\V_1~~ \V_2], \qquad \V_1\in \R^{n\times n_{\text{AS}}}, ~~\V_2\in\R^{n\times(n-n_{\text{AS}})}.
\label{eq:decomp}
\end{equation}
The space spanned by $\V_1$ columns is called the \textit{active subspace} of dimension $n_{\text{AS}} < n$, whereas the \textit{inactive subspace} is
defined as the range of the remaining eigenvectors in $\V_2$. Once we have
defined these spaces, the input $\mupar\in\R^n$ can be reduced to a low-dimensional
vector $\tilde{\mupar}_1\in\R^{n_{\text{AS}}}$ using $\V_1$ as projection map. To be more precise, any
$\mupar\in\R^n$ can be expressed in this way using the decomposition in
Eq.~\eqref{eq:decomp} and the properties of $\V$:
\begin{equation}
\mupar = \V\V^T\mupar = \V_1\V_1^T\mupar + \V_2\V_2^T\mupar = \V_1\tilde{\mupar}_1 + \V_2\tilde{\mupar}_2,
\end{equation}
where the two new variables $\tilde{\mupar}_1$ and $\tilde{\mupar}_2$ are the
\textit{active} and \textit{inactive variable} respectively:
\begin{equation}\label{eq:active_var}
\tilde{\mupar}_1 = \V_1^T\mupar \in \R^{n_{\text{AS}}}, \qquad ~ \tilde{\mupar}_2 = \V_2^T\mupar \in \R^{n-n_{\text{AS}}}.
\end{equation}
For the actual computations of the AS we have used the open source Python
package called ATHENA~\cite{athena2020}.

%% file: sections/POD.tex
In this section, we are going to describe the Proper Orthogonal Decomposition (POD) approach of reduce order modeling~\cite{rozza2015book} by decreasing the number of degrees of freedom of a parametric system. In particular we are focusing on the POD with interpolation (PODI) method \cite{bui2003proper, bui2004aerodynamic, ly2001modeling, DemoTezzeleGustinLaviniRozza2018, DemoTezzeleMolaRozza2018, DemoTezzeleMolaRozza2019}.\\
Let $\mathbf{S} = [\upar_1\dots \upar_{n_S}]$ be the matrix of snapshots, i.e. the full order system outputs $\upar_i\in\R^N$. Once these solutions are collected we aim to describe them as a linear combination of few main structures, the POD modes, and thus project them onto a low dimensional space spanned by these modes. In order to calculate the POD modes, we need to compute the singular value decomposition (SVD) of the snapshots matrix $\mathbf{S}$:
\begin{equation}\label{eq:svd_pod}
\mathbf{S} = \mathbf{\Psi}\mathbf{\Sigma}\mathbf{\Theta}^T,    
\end{equation}
where the left-singular vectors, i.e. the columns of the unitary matrix $\mathbf{\Psi}$, are the POD modes, and the diagonal matrix $\mathbf{\Sigma}$ contains the corresponding singular values in decreasing order. Therefore, by selecting the first modes we are retaining only the most energetic ones and we can construct a reduced space into which we project the high-fidelity solutions. Hence we obtain:
\begin{equation}\label{eq:pod_red}
\mathbf{S}^{\text{POD}}=\mathbf{\Psi}^T_{N_{\text{POD}}}\mathbf{S}.    
\end{equation}
where $\mathbf{\Psi}_{N_{\text{POD}}}$ is the matrix containing only the first $N_{\text{POD}}$ modes and the columns of $\mathbf{S}^{\textrm{POD}}$ represent the reduced snapshot $\tilde{\upar}_i\in\R^{N_{\text{POD}}}$.

%% file: sections/PCE.tex
The Polynomial Chaos Expansion (PCE) theory was initially proposed by Wiener in~\cite{wiener}, showing that a real-valued random variable~$X:\R^R\to\R$ can be decomposed
in the following way:
\begin{equation}\label{eq:PCE}
X(\bm{\xi}) = \sum_{j=0}^{\infty} c_j \phipar_j(\bm{\xi}), 
\end{equation}
i.e. as an infinite sum of orthogonal polynomials weighted by unknown
deterministic coefficients $c_j$ \cite{janya2017framework}. The vector $\bm{\xi} = (\xi^1, \dots, \xi^R)$
represents the multi-dimensional random vector, where each element is
associated with uncertain input parameters, while $\phipar_j(\bm{\xi})$ are multivariate
orthogonal polynomials, that can be decomposed into products of one-dimensional
orthogonal polynomials with different variables.\\
We can approximate the infinite sum in Eq.~\eqref{eq:PCE} by truncating it at the $(P+1)$-th term, such that:
\begin{equation}\label{eq:PCE1}
X(\bm{\xi}) \approx \sum_{j=0}^{P} c_j \phipar_j(\bm{\xi}),
\end{equation}
with the number of unknown coefficients in this summation given by $P+1 =
\frac{(p+R)!}{p!R!}$~\cite{ghanem2003stochastic}, where $p$ is the degree of the polynomial we are
considering in the $R$-dimensional space.\\
When the parameters $\xi^1, \dots, \xi^R$ are independent, $ \phipar_j(\bm{\xi})$
can be decomposed into products of one-dimensional functions:
\begin{equation}
\phipar_j(\bm{\xi}) = \phipar_j(\xi^1, \dots, \xi^R) = \prod_{k=1}^R \phi_k^{d_k}(\xi^k), ~~~~j=0,\dots,P,~~~~d_k=0,\dots,p, ~~~~ s.t. ~ \sum_{k=1}^{R}d_k\le p.
\end{equation}
In order to determine the PCE, we need to find out the polynomial chaos expansion coefficients $c_j$ for $j = 0, \dots, P$, and the one-dimensional orthogonal polynomial $\phi_k^{d_k},~ k=1,\dots,R$, of degree~$d_k$.\\
Based on the work of Askey and Wilson \cite{Askey}, we can provide the orthogonal polynomials for different distribution. One of the possible choices is respresented by the Gaussian distribution with the related Hermite polynomials.\\
The estimation of the coefficients of PCE can be carried out in different ways \cite{sudret}: following a projection method based on the orthogonality of the polynomials or following a regression method, that is the one we are going to describe.\\
In order to determine the coefficients $c_j$, we need to solve a minimization problem:
\begin{equation}\label{eq:coeff}
\mathbf{c} = \argmin_{\mathbf{c}^*\in\R^P} \frac{1}{N_{\text{PCE}}}\sum_{i=1}^{N_{\text{PCE}}} \left( \hat{X}  - \sum_{j=0}^{P} c^*_j \phipar_j(\bm{\xi}_{i})\right),
\end{equation}
where $N_{\text{PCE}}$ indicates the total number of realizations of the input vector we are considering, whereas $\hat{X}$ represents the real output of the model. In order to solve equation \cref{eq:coeff} we need to consider the following matrix:
\begin{equation}
\Phipar = \begin{pmatrix}
\phipar_0(\bm{\xi}_{1}) & \phipar_1(\bm{\xi}_{1}) & \cdots & \phipar_{P}(\bm{\xi}_{1}) \\
\phipar_0(\bm{\xi}_{2}) & \phipar_1(\bm{\xi}_{2}) & \cdots & \phipar_{P}(\bm{\xi}_{2}) \\
\vdots & \vdots & \ddots & \vdots\\
\phipar_0(\bm{\xi}_{N_{\text{PCE}}}) & \phipar_1(\bm{\xi}_{N_{\text{PCE}}}) & \cdots & \phipar_{P}(\bm{\xi}_{N_{\text{PCE}}}) 
\end{pmatrix}.
\end{equation}
Thus, the solution of equation \cref{eq:coeff} is computed by a least-square optimization:
\begin{equation}
\mathbf{c} = (\Phipar^T\Phipar)^{-1} \Phipar^T \hat{X},
\end{equation}
If the matrix $\Phipar^T\Phipar$ is ill-conditioned, as it may happen, the singular value decomposition method should be employed.

%% file: sections/ANN.tex
\begin{figure}[htbp]
    \centering
    \tikzset{%
      every neuron input/.style={
        circle,
        draw,
        minimum size=0.7cm,
        fill=green!50
      },
      every neuron input2/.style={
        circle,
        draw,
        minimum size=0.7cm,
        fill=yellow!50
      },
       every neuron hidden/.style={
        circle,
        draw,
        minimum size=0.7cm,
        fill=blue!50
      },
       every neuron output/.style={
        circle,
        draw,
        minimum size=0.7cm,
        fill=red!50
      },
      neuron missing/.style={
        draw=none,
        fill=none,
        scale=4,
        text height=0.333cm,
        execute at begin node=\color{black}$\vdots$
      },
       layer missing/.style={
        draw=none,
        scale=4,
        text height=0.333cm,
        execute at begin node=\color{black}$\dots$
      },
    }
    \begin{tikzpicture}[x=1.5cm, y=1.5cm, >=stealth, font=\small, scale=0.9]
    \foreach \i [count=\y] in {1, 2, missing, 3}
    \node [every neuron input2/.try, neuron \i/.try] (input-\i) at (0,2.5-\y) {};
    \foreach \m [count=\y] in {1, 2, missing,3}
      \node [every neuron hidden/.try, neuron \m/.try ] (hidden1-\m) at (2,2.5-\y) {};
    \foreach \m [count=\y] in {1, missing, 2, 3}
      \node [every neuron hidden/.try, neuron \m/.try ] (hidden2-\m) at (4,2.5-\y) {};
    \foreach \m [count=\y] in {1,missing,2}
      \node [every neuron output/.try, neuron \m/.try ] (output-\m) at (6,1.9-\y) {};
    \draw [<-] (input-3) -- ++(-1,0)
        node [above, midway] {$\tilde{x}^{n_{\text{in}}}$};
    \draw [<-] (input-1) -- ++(-1,0)
        node [above, midway] {$\tilde{x}^1$};
    \draw [<-] (input-2) -- ++(-1,0)
        node [above, midway] {$\tilde{x}^2$};
    \foreach \l [count=\i] in {1,{n_{\text{out}}}}
      \draw [->] (output-\i) -- ++(1,0)
        node [above, midway] {$h^{\l}$};
    \foreach \i in {1,...,3}
      \foreach \j in {1,...,3}
        \draw [->] (input-\i) -- (hidden1-\j);
    \foreach \i in {1,...,3}
      \foreach \j in {1,...,3}
        \draw [->] (hidden1-\i) -- (hidden2-\j);
    \foreach \i in {1,...,3}
      \foreach \j in {1,...,2}
        \draw [->] (hidden2-\i) -- (output-\j);
    \foreach \l [count=\x from 0] in {Input layer}
    \node [align=center, above] at (\x*2,2) {\l};
     \node [align=center, above] at (2,2) {Hidden layer 1};
    \node [align=center, above] at (4,2) {Hidden layer 2};
    \node [align=center, above] at (6,2) {Output layer};
    \end{tikzpicture}
    \caption{Schematic structure of a Feedforward Neural Network with 2 hidden layers.}
    \label{fig:FNN}
\end{figure}
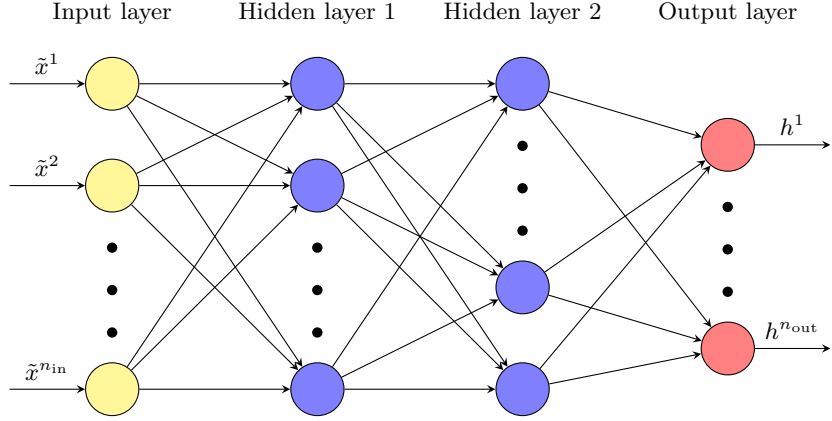

A Feedforward Neural Network (FNN), also called \emph{multilayer perceptron}, is a popular neural network model, usually employed for function regression \cite{fine2006feedforward}. As depicted in~\cref{fig:FNN}, it mainly consists of an input layer, an output layer and a certain number of hidden layers\footnotemark \footnotetext{A priori there is not a right number of hidden layers to use: it depends on the fields of application of your net and on the problem in exam \cite{trenn2008}}, where the processing units composing them are called \textit{neurons}. Each neuron is defined by a weight vector, that characterizes the strength of the connection with the neurons in the next layer.\\
More technically speaking, let $\tilde{\x}\in \R^{n_{\text{in}}}$ be the input vector and $M$ the total number of hidden layers of the FNN. The output vector $\h \in \R^{n_{\text{out}}}$ is obtained through the application of an \textit{activation function} to the weighted sum of all the inputs arriving to it. The activation function is used in order to introduce non-linearity in the network and some common choices are represented by the ReLU function, the sigmoid, the logistic function or the radial activation functions, \cite{Goodfellow}. \\
In order to understand better how to get the general formula \eqref{eq:ANN}, we start by considering a simple FNN with a single output and one hidden layer. In this case the final output is given by:
\begin{equation}
\h = \sigma(\sum_{i=1}^{n_{\text{in}}} w_i \tilde{x}_i +b_i)
\end{equation}
where $\sigma$ is the activation function, $W = \{w_i\}_{i=1}^{n_{\text{in}}}$ represents the weights of the net and $b$ the bias\footnotemark\footnotetext{For simplicity the bias is put to zero in the following discussion.}. Thus, if we consider $M$ layers the final output can be seen as a weighted sum of its input followed by the activation function, where each input can be rewritten in the same way:
\begin{align}
\label{eq:ANN}
\begin{split}
h^j &= \sigma\left(\sum_{i=1}^{n_{M}} w^{M+1}_{ji} \tilde{x}^{M,i}\right) = \sigma\left(\sum_{i=1}^{n_{M}} w^{M+1}_{ji} \left(\sigma\left(\sum_{q=1}^{n_{M-1}} w^M_{iq} \tilde{x}^{M-1,q}\right)\right)\right)=\dots=\\
&= \sigma\left(\sum_{i=1}^{n_{M}} w^{M+1}_{ji} \left(\sigma\left(\sum_{q=1}^{n_{M-1}} w^M_{iq} \left(\sigma\left(\dots\left(\sigma\left(\sum_{k=1}^{n_{in}} w^1_{sk} \tilde{x}^k\right)\right)\right)\right)\right)\right)\right),~~~~~~~ j = 1, \dots, n_{\text{out}},
\end{split}
\end{align}
where $n_m$, $m=1,\dots,M$, represents the number of neurons in layer $m$, whereas $n_{\text{in}}$ and $n_{\text{out}}$ are the neurons in the input and output layers respectively, $W^m= (w_{ki}^m)_{ki},~ k=1,\dots,n_m, ~ i=1,\dots,n_{m-1}$ indicates the weight matrix related to layer $m$. Note that the first number in any weight’s subscript matches the index of the neuron in the next layer and the second number matches the index of the neuron in the previous layer.\\
One of the main characteristics of an FNN is the ability to learn from observational data during the so-called \textit{training process}. In this phase the net is acquiring knowledge from our dataset by minimizing the loss function\footnotemark\footnotetext{There exists several types of loss functions that are used in this context, such as the Cross-Entropy Loss, the Euclidean Loss and the Hinge Loss. Also in this case, the correct choice depends on the problem in exam~\cite{alzubaidi2021review, Goodfellow}} $\mathcal{L}$:
\begin{equation}\label{eq:min_pb_19}
\min_W\left\{\frac{1}{n_{\text{out}}}\sum_{i=1}^{n_{\text{out}}} \mathcal{L}(h^i,\hat{h}^i)   \right\},
\end{equation}
where $\h= \{h^j \}_{j=0}^{n_{\text{out}}}$ represents the expected output and $\hat{\h}= \hat{\h}(\tilde{\x}; W)= \{\hat{h}^j(\tilde{\x}; W) \}_{j=0}^{n_{\text{out}}}$ is the prediction made by our FNN. In order to solve this minimization problem the \textit{Backpropagation algorithm} \cite{rojas1996backpropagation} is employed. Therefore, the model's parameters are optimized by adjusting network’s weights as follows:
\begin{equation}
w_{ki}^{m,(t)} = w_{ki}^{m, (t-1)} -\epsilon \frac{d\mathcal{L}}{dw_{ki}^m},  
\end{equation}
where $\epsilon$ is the \textit{learning rate}, appropriately chosen according to the problem in exam, and $t$ represents the training epoch, i.e. a complete repetition of the parameter update that involves the complete training dataset at one time. The computation of the gradients is then performed by exploiting the chain rule.

%% file: sections/ASNN.tex
\begin{figure}
    \centering
    \includegraphics[width=.8\textwidth]{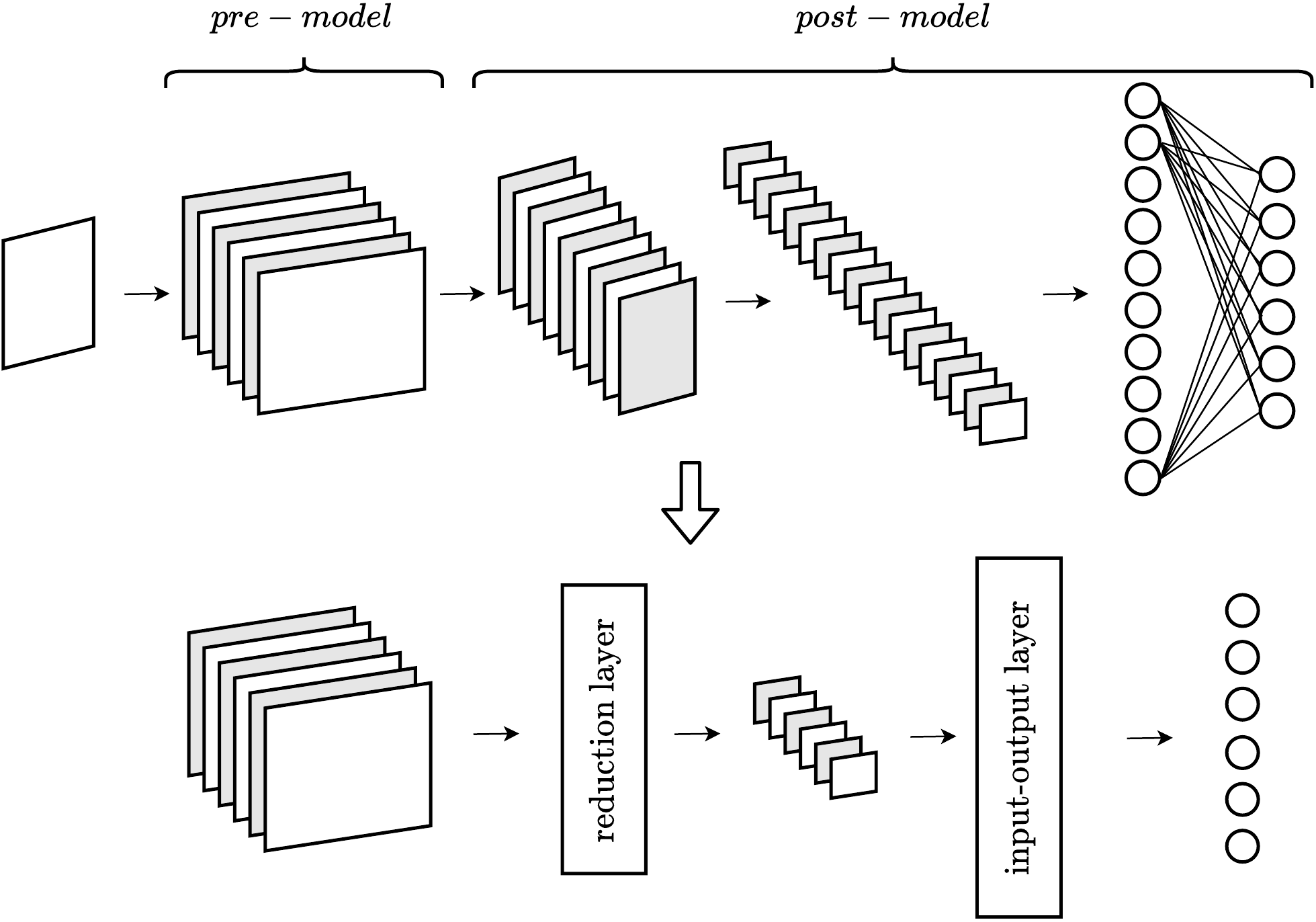}
    \caption{Graphical representation of the reduction method proposed for a CNN.}
    \label{fig:cnn_red}
\end{figure}

We provide in this section the rigorous description of the proposed framework (summarized in~\cref{fig:cnn_red} and~\cref{alg:cnn_red}), which has the final goal of reducing in dimensionality a generic artificial neural network (ANN). Indeed, the only assumption we make regarding the original network is that it is composed by $L$ layers.
\paragraph{Network splitting}
At the beginning, the original network $\fullnn: \R^{n_0} \to \R^{n_L}$ is split in two different parts such that the first $l$ layers constitutes the \emph{pre-model} while the last $L-l$ layers form the so-called \emph{post-model}. Describing the network as composition of functions $\fullnn \equiv f_L \circ f_{L-1} \circ \dots \circ f_1$, we can formally define the pre- and the post-model as:
\begin{equation}
\fullnn_{\text{pre}}^l =  f_l \circ f_{l-1} \circ \dots \circ f_1,\qquad \qquad \fullnn_{\text{post}}^l =f_L \circ f_{L-1} \circ \dots \circ f_l,
\end{equation}
where the functions $f_j: \R^{n_{j-1}} \to \R^{n_j}$ for $j=1,\dots,L$, represent the different layers of the network --- e.g. convolutional, fully connected, batch-normalization, ReLU, pooling layers.
The original model can then be rewritten as 
\begin{equation}
\fullnn(\x^0) = \fullnn^l_{\text{post}}(\fullnn^l_{\text{pre}}(\x^0)),
\end{equation}
for any $1\le l < L$.
The reduction of the network effectively happens approximating the post-model, which means that the pre-model is actually copied from the original network to the reduced one.
Before proceeding with the algorithm to explain how the post-model is approximated, we specify that the index $l$, denoting the \textit{cut-off layer}, is the only parameter of this initial step, and it plays an important role in the final outcome. This index indeed defines how many layers of the original network are kept in the reduced architecture, controlling in few words how many information of the original network we are discarding.
It should be then chosen empirically based on considerations about the network and the dataset at hand, balancing the final accuracy and the compression ratio.

\begin{algorithm}
\caption{Pseudo-code for the construction of the reduce artificial neural network}\label{alg:cnn_red}
\textbf{Inputs:} \begin{itemize}
\item a dataset with $N_{\text{train}}$ input samples $\mathcal{D}_0=\{\x^0_{j}\}_{j=1}^{N_{\text{train}}}$, 
\item an artificial neural network $\fullnn$,
\item $\{\y_j\}_{j=1}^{N_{\text{train}}}$ real output of the $\fullnn$, 
\item reduced dimension $r$, 
\item index of the cut-off layer $l$
\end{itemize}
\begin{algorithmic}[1]
\State{$\fullnn_{\text{pre}}^l, \fullnn_{\text{post}}^l =$ splitting\_net$(\fullnn, l)$}
\State{$\x^l=\fullnn^l_{\text{pre}}(\x^0)$}
\State{$\z =$ reduce$(\x^l, r)$}
\State{$\hat{\y} =$ input\_output\_map$(\z, \y)$}
\State{Training of the constructed reduced net}
\end{algorithmic}
\textbf{Output:} Reduced Net $\fullnn^{\text{red}}$
\end{algorithm}

\paragraph{Dimensionality reduction}
As introduced previously, we aim to project the output  $\x^l$ of the pre-model onto a low-dimensional space using reduction techniques as:  
\begin{itemize}
\item \textbf{Active Subspaces}: as described in~\cref{sec:AS} and in \cite{cui2020active}, we consider a function $g_l$ defined by:
\begin{equation}
g_l(\x^l) = \text{loss} (\fullnn^l_{\text{post}}(\x^l)),
\end{equation} 
in order to extract the most important directions and determine the projection matrix used to reduce the pre-model output.
\item \textbf{Proper Orthogonal Decomposition}: as discussed in~\cref{sec:POD}, the SVD decomposition~\cref{eq:svd_pod} is exploited to compute the projection matrix $\mathbf{\Psi}_r$ and thus the reduced solution
\begin{equation}
\z= \mathbf{\Psi}^T_r\x^l.   
\end{equation}
\end{itemize}

\paragraph{Input-Output mapping}
The last part of the reduced net is dedicated to the classification of the output coming from the reduction layer. In order to do so, we have employed two different techniques:
\begin{itemize}
\item the \textbf{Polynomial Chaos Expansion} introduced in~\cref{sec:PCE}. As described in equation~\cref{eq:PCE1}, the final output of the network $\y=\fullnn(\x^0)\in\R^{n_L}$, i.e. the true response of the model, can be approximated in the following way:
\begin{equation}
\hat{\y}\approx \sum_{\lvert \alphapar \rvert=0}^{p}\cpar_{\alphapar}\phipar_{\alphapar}(\z), \qquad \lvert \alphapar \rvert=\alpha_1+\dots+\alpha_r,
\end{equation}
where $\phipar_{\alphapar}(\z)$ are the multivariate polynomial functions chosen based on the probability density function $\rho$ associated with $\z$. Therefore, the estimation of coefficients $\cpar_{\alpha}$ is carried out by solving the minimization problem~\cref{eq:coeff}:
\begin{equation}
\min_{c_{\alpha}}\frac{1}{N_{\text{train}}}\sum_{j=1}^{N_{\text{train}}}\left\lVert\y_j-\sum_{\lvert \alphapar \rvert=0}^{p}\cpar_{\alphapar}\phipar_{\alphapar}(\z_j)\right\rVert^2.
\end{equation}

\item a \textbf{Feedforward Neural Network} described in~\cref{sec:ANN}. In this case, the output of the reduction layer $\z$ coincides with the network input and by using Equation~\cref{eq:ANN} we obtain that the final output $\hat{\y}$ of the reduced net\footnotemark\footnotetext{Note that in this case the number of hidden layers is set to $1$ since, as discussed in~\cref{sec:results}, we notice that one hidden layer is enough to gain a good level of accuracy (see for example~\cref{ann_cifar}).} is determined by:
\begin{equation}
\hat{y}^j = \sum_{i=1}^{n_1}w_{ji}^{2}z^{1,i} = \sum_{i=1}^{n_1}w_{ji}^{2} \sigma\left(\sum_{m=1}^{r} w_{im}^{1} z^m\right), ~~~~ j = 1,\dots,n_{\text{out}},
\end{equation}
where $n_{\text{out}}$ corresponds to the number of categories that compose the dataset in exam, and $\sigma$ is the \textit{Softplus} function:
\begin{equation}
    \textrm{Softplus}(\x) = \frac{1}{\beta}\log(1+\exp(\beta\x)).
\end{equation}
\end{itemize}

\subsection{Training phase}
Once the reduced version of the network in exam is constructed, we need to train it. Following~\cite{cui2020active}, for the training phase of the reduced ANN the technique of \textit{knowledge distillation}~\cite{hinton}, is used. A knowledge distillation framework contains a large pre-trained \textit{teacher model}, our full network, and a small \textit{student model}, in our case $\fullnn^{\textrm{red}}$. Therefore, the main goal is to train efficiently the student network under the guide of the teacher network in order to gain a comparable or even superior performance.\\
Let $\y$ be a vector of \textit{logits}, i.e. the output of last layer in a deep neural network. The probability $p_i$ that the input belongs to the $i$-th class is given by the softmax function
\begin{equation}
p_i = \frac{exp(y^i)}{\sum_{j=0}^{n_{\text{class}}} exp(y^j)}.
\end{equation}
As described in~\cite{hinton}, a temperature factor $T$ need to be introduced in order to control the importance of each target 
\begin{equation}
p_i = \frac{exp(y^i/T)}{\sum_{j=0}^{n_{\text{class}}} exp(y^j/T)},
\end{equation}
where if $T\to\infty$ all classes have the same probability, whereas if $T\to 0$ the targets $p_i$ become one-hot labels.\\
First of all, we need to define the \textit{distillation loss}, that matches the logits between the teacher model and the student model. As done in~\cite{cui2020active}, the \textit{response-based knowledge} is used to transfer the knowledge from the teacher to the student by mimicking the final prediction of the full net. Therefore, in this case the distillation loss~\cite{KD, hinton} is given by:
\begin{equation}\label{eq:LD}
L_D(p(\y_t, T), p(\y_s, T)) = \mathcal{L}_{\text{KL}}(p(\y_t,T), p(\y_s, T)),
\end{equation}
where $\y_t$ and $\y_s$ indicate the logits of the teacher and student networks, respectively, whereas $\mathcal{L}_{\text{KL}}$ represents the Kullback-Leibler (KL) divergence loss \cite{kim2021comparing}:
\begin{equation}
\mathcal{L}_{\text{KL}}((p(\y_s,T), p(\y_t, T)) = T^2 \sum_j p_j(y_t^j, T)\log \frac{p_j(y_t^j, T)}{p_j(y_s^j, T)}.
\end{equation}
The \textit{student loss} is defined as the cross-entropy loss between the ground truth label and the logits of the student network \cite{KD}:
\begin{equation}\label{eq:LS}
L_S(\y, p(\y_s,T)) = \mathcal{L}_{\text{CE}}(\hat{\y}, p(\y_s,T)),
\end{equation}
where $\hat{\y}$ is a ground truth vector, characterized by having only the component corresponding to the ground truth label on the training sample set to 1 and the others are 0. Then, $\mathcal{L}_{\text{CE}}$ represents the cross entropy loss
\begin{equation}
\mathcal{L}_{\text{CE}} (\hat{\y}, p(\y_s,T))=\sum_i -\hat{y}^i \log(p_i(y_s^i, T)).
\end{equation}
As can be observed, both losses, \cref{eq:LD} and \cref{eq:LS}, use the same logits of the student model but with different temperatures: $T=\tau>1$ in the distillation loss, and $T=1$ in the student loss. Finally, the final loss is a weighted sum between the distillation loss and the student loss:
\begin{equation}
L(\x^0, W) = \lambda L_D(p(\y_t, T=\tau), p(\y_s, T=\tau)) + (1-\lambda) L_S(\hat{\y}, p(\y_s , T=1)),
\end{equation}
where $\lambda$ is the regularization parameter, $\x^0$ is an input vector of the training set, and $W$ are the parameters of the student model.

%% file: sections/results.tex
In this section we present a comparison between the results obtained with the different reduction methods proposed in terms of final accuracy, memory allocation, and speed of the procedure. 

\subsection{VGG-16}
\begin{figure*}[ht]
\captionsetup[subfigure]{labelformat=empty}
\centering
\includegraphics[width=0.96\textwidth]{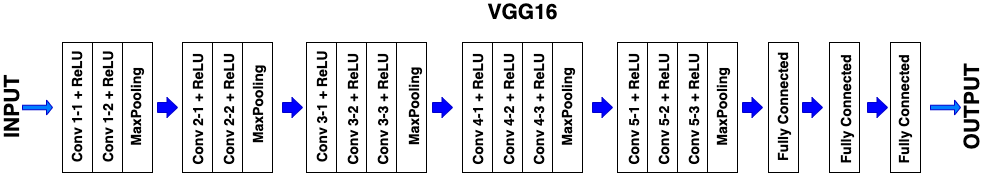}
\caption{Graphical representation of VGG-16 architecture.}
\label{fig:VGG16}
\end{figure*}

As test network, we use a Convolutional Neural Network (CNN), a class of ANN commonly applied for the problem of image recognition~\cite{rawat2017deep}. Over the last 10 years, several CNN architectures have been presented~\cite{ajit2020review, alzubaidi2021review} to tackle this problem, e.g. AlexNet, ResNet, Inception, VGGNet. As starting point to test our methods, we employ one of the VGG network architecture: VGG-16~\cite{vgg}. As can be seen in~\cref{fig:VGG16}, the architecture is composed by:
\begin{itemize}
\item 13 \textit{convolutional blocks}: each block is made of a convolutional layer followed by a non-linear layer, i.e. the application of the activation function, in this case ReLU is used.
\item 5 \textit{max-pooling layers}, 
\item 3 \textit{fully-connected layers}.
\end{itemize}
The total number of layers having tunable parameters is 16 of which 13 are the convolutional layers and 3 the fully connected layers. For this reason the name given to this ConvNet is VGG-16.

\subsection{Dataset} For training and testing our net we use:
\begin{itemize}
    \item \textbf{CIFAR-10 dataset}~\cite{cifar10}, a computer-vision dataset used for object recognition. It consists of $60000$ $32\times 32$ colour images divided in $10$ non-overlapping classes: airplane, automobile, bird, cat, deer, dog, frog, horse, ship, and truck.
    \item \textbf{Custom dataset}, composed of $3448$ $32\times 32$ colour images organized in $4$ classes: 3 non-overlapping classes and a mixed one, characterized by pictures with objects of different categories present at the same time.
\end{itemize}

\subsection{Software}
In order to implement and construct the reduced version of the convolutional neural network presented in the previous sections, we employed PyTorch~\cite{pytorch} as development environment. We then used the open-source Python library SciPy~\cite{SciPy} for scientific computing and the open source Python package ATHENA~\cite{athena2020} for the actual computation of the active subspaces.

\subsection{Results} 
We now present the results of the reduced net constructed starting from VGG-16 and based on CIFAR-10 and our custom dataset. First of all, the original network VGG-16 has been trained on each of the different dataset presented for $60$ epochs\footnotemark.\footnotetext{We have chosen $60$ (and then 10 for the reduced net) to be the number of epochs for the training phase as a trade-off between the final accuracy and the time needed. For this reason, we kept the same value in the two different cases we are considering in order to have a fair comparison. } From \cref{cifar:res} and \cref{custom:res}, it can be seen that at the end of this training VGG-16 gain a good accuracy: $77.98\%$ for the CIFAR-10 and $95.65\%$ for the custom dataset.\\
We report the results obtained with different reduced versions of VGG-16 constructed following the steps of~\cref{alg:cnn_red} and using three cut-off layers\footnotemark~$l$:~$5, 6$, and $7$, as done in~\cite{cui2020active}.\footnotetext{In~\cite{cui2020active} and in the correspondent implementation they refer to indices $5,6$ and $7$, i.e. the indices of the convolutional layers in a list where only convolutional and linear layers are taken into consideration as possible cut-off layers. Thus, if we take into account the whole net with all the different layers, we are considering layers $11, 13$, and $16$.} We remark that in the case of dimensionality reduction with the Active Subspaces technique, we employed the \textit{Frequent Direction method}~\cite{FrequentDir} implemented inside ATHENA to compute the AS. The value chosen for the parameter $r$, i.e. the dimension of the reduced space, is $50$ both for AS and for POD in analogy with~\cite{cui2020active}, where considerations on the structural analysis of VGG-16 can be found.\\
In the case a FNN is used to classify our pictures, we trained it for $500$ epochs with our dataset before re-training the whole reduced net. In~\cref{ann_cifar}, we summarize the results obtained by training the reduced net using different FNN architectures, i.e. a different number of hidden layers and also of hidden neurons, which are kept constant in each hidden layer of the net.  In particular we are making a comparison between the storage needed for the FNN and the accuracy of the reduced net in exam at epoch $0$, i.e. after its initialization, and at epoch $10$, i.e. after the re-training of the whole reduced net. Thus, it can be seen that increasing the number of hidden layers and also of hidden neurons is not giving a gain in accuracy. For this reason, based on considerations about the final accuracy and the allocation in memory of the FNN (see~\cref{ann_cifar} for details), we decided to use the following architectures: 
\begin{itemize}
    \item \textbf{CIFAR10}: FNN with $50$ input neurons, $10$ output neurons, and one hidden layer with $20$ hidden neurons.
    \item \textbf{Custom Dataset}: FNN with $50$ input neurons, $4$ output neurons, and one hidden layer with $10$ hidden neurons.
\end{itemize}

\newcommand{\ccc}{\cellcolor{gray!10}}  
\begin{table}[htbp]
\small
\centering
\caption{Results obtained for the reduced net POD+FNN (7) trained on CIFAR10 with different structures for the FNN.}
\begin{tabular}{cc|ccccc}
 \toprule
 \multicolumn{3}{c}{} &\multicolumn{4}{c}{\bf Hidden layers} \\
 \multicolumn{3}{c}{} &\bf 1 & \bf 2&\bf  3& \bf 4 \\
\midrule
 \multirow{4}{*}{\rotatebox{90}{\bf Hidden neurons~~~~~~~~~~~}} & 
 \multirow{3}{*}{\bf 10} &  \ccc Epoch 0  & \ccc 81.39\% &6\ccc 7.92\% &\ccc 75.52\%  &\ccc  81.57\% \\
 & & Epoch 10& 87.89\% &87.59\% &87.46\% & 87.26\% \\
 & &\ccc  Storage FNN (Mb) &\ccc  0.0024 &\ccc 0.0028 &\ccc 0.0032 &\ccc  0.0036
 \\\cmidrule{2-7}
 &\multirow{3}{*}{\bf 20} & Epoch 0  & 80.17\% &80.05\% & 79.97\%  & 78.28\% \\
 & &\ccc  Epoch 10&\ccc  87.45\%&\ccc  87.13\%&\ccc  87.42\%&\ccc  86.68\%\\
 & & Storage FNN (Mb) & 0.0047 &0.0063 &0.0079 & 0.0095 \\\cmidrule{2-7}
&\multirow{3}{*}{\bf 30} &\ccc  Epoch 0  &\ccc  77.57\% &\ccc  80.36\%&\ccc  80.43\% &\ccc  76.26\% \\
 & & Epoch 10& 86.92\% & 86.25\%&86.30\% & 85.25\%\\
 & &\ccc  Storage FNN (Mb) &\ccc  0.0070 &\ccc 0.0106 &\ccc 0.0141 &\ccc  0.0177 \\\cmidrule{2-7}
 &\multirow{3}{*}{ \bf 40} & Epoch 0  & 71.24\%  &70.38\% & 69.31\%  & 68.15\% \\
 & &\ccc  Epoch 10&\ccc  85.04\%&\ccc  84.60\%&\ccc  84.18\% &\ccc  83.64\%\\
 & & Storage FNN (Mb) & 0.0093 &0.0156 &0.0219 & 0.0281 \\
\bottomrule
\end{tabular}
\label{ann_cifar}
\end{table}
Hence, after these steps, the reduced net has been re-trained for $10$ epochs on each dataset. The results obtained are summarized in~\cref{cifar:res} and~\cref{custom:res} showing a comparison between the different reduced nets in terms of accuracy (before and after the final training), memory storage, and time needed for the initialization and the training of the reduced net. As discussed previously for each reduced net, i.e. AS+PCE, AS+FNN, POD+FNN, we proposed in the tables the results achieved using three different cut-off layers: 5,6, and 7.

\begin{table}
\centering
\footnotesize
\rowcolors{3}{}{gray!10}
\caption{Results obtained with CIFAR10 dataset.}
\begin{tabular}{c|cc|ccc|cc}
 \toprule
\bf Network & \multicolumn{2}{c|}{\bf Accuracy} & \multicolumn{3}{c|}{\bf Storage (Mb)} &  \multicolumn{2}{c}{\bf Time} \\
 \midrule
 \rowcolor{gray!10}

 VGG-16   & \multicolumn{2}{c|}{77.98\%} & \multicolumn{3}{c|}{56.15} & \multicolumn{2}{c}{46 h} \\
 \midrule
& \bf Epoch 0 & \bf Epoch 10   & \bf Pre-M & \bf  AS/POD & \bf PCE/FNN & \bf Init & \bf Train\\
 \midrule
 AS+PCE (5)&  13.52\%  & 82.01\% & 2.12 & 3.12 & 0.05 & 43 min & 4.5 h\\
 AS+FNN (5)& 33.06\%  & 80.43\% & 2.12 & 3.12  & 0.0047  & 5 h & 4.5 h\\
 POD+FNN (5)& 62.16\%   & 80.24\% & 2.12 &3.12  &0.0047  &79 min & 5 h\\
 AS+PCE (6)&  14.42\%  & 84.69\% & 4.37 & 3.12 & 0.05 &49 min &5.5 h\\
 AS+FNN (6)& 33.76\%   & 82.13\% & 4.37 & 3.12  & 0.0047  & 5 h & 4.5 h\\
 POD+FNN (6)&  63.84\%  &83.93\%  & 4.37 & 3.12 & 0.0047 & 83 min & 5 h\\
 AS+PCE (7)& 4.25\%   & 85.60\% & 6.62 & 0.78 & 0.05 &35 min & 5.5 h\\
 AS+FNN (7)& 75.66\%   & 86.03\% & 6.62 & 0.78 & 0.0047 & 1.5 h & 5 h\\
 POD+FNN (7)& 80.17\%   & 87.45\% &  6.62 & 0.78 & 0.0047  & 12 min & 5 h\\
 \bottomrule
\end{tabular}
\label{cifar:res}
\end{table}

\begin{table}
\footnotesize
\rowcolors{3}{}{gray!10}
\centering
\caption{Results obtained with a custom dataset.}
\begin{tabular}{c|cc|ccc|cc}
 \toprule
\bf Network & \multicolumn{2}{c|}{\bf Accuracy} & \multicolumn{3}{c|}{\bf Storage (Mb)} & \multicolumn{2}{c}{\bf Time} \\
 \midrule
\rowcolor{gray!10}
 VGG-16   & \multicolumn{2}{c|}{95.65\%} & \multicolumn{3}{c|}{56.14}  &\multicolumn{2}{c}{22 min} \\
 \midrule
& \bf Epoch 0 & \bf Epoch 10   & \bf Pre-M & \bf AS/POD & \bf PCE/FNN  & \bf Init & \bf Train\\
 \midrule
 AS+PCE (5)&  29.03\%  &95.21\%  & 2.12 & 3.12 & 0.02  & 2 min & 10 min\\
 AS+FNN (5)&  94.63\%  & 94.92\% & 2.12 & 3.12&  0.0021 & 12.5 min & 12 min\\
 POD+FNN (5)& 96.52\%   & 96.66\%  & 2.12 & 3.12 & 0.0021 & 28 sec & 11.5 min\\
 AS+PCE (6)& 29.75\%   &  95.79\%  & 4.37 & 3.12& 0.02& 2.5 min & 10 min\\
 AS+FNN (6)&  94.92\%  & 95.36\% & 4.37 & 3.12 & 0.0021 & 12.5 min & 12.5 min\\
 POD+FNN (6)&  96.23\%  & 96.37\% & 4.37  & 3.12  &0.0021 &  33 sec& 13 min\\
 AS+PCE (7)&  28.59\%  & 94.05\% & 6.62 & 0.78 & 0.02 & 1.5 min & 11 min\\
 AS+FNN (7)&  94.34\%  & 94.63\% & 6.62 & 0.78 & 0.0021 & 4.5 min & 13 min\\
 POD+FNN (7)& 96.37\%   &96.52\%  &6.62  & 0.78  & 0.0021 & 33 sec & 14 min\\
 \bottomrule
\end{tabular}
\label{custom:res}
\end{table}
Information on memory allocation are important since in our context (in particular the custom dataset) we need to include a convolutional neural network in an embedded system with particular constraints on the storage. In both~\cref{cifar:res} and~\cref{custom:res} it can be seen that the allocation required for the created reduced nets is decreased with respect to that of the original VGG-16. In fact, the checkpoint file\footnotemark\footnotetext{Note that in both cases (CIFAR10 and custom dataset) the checkpoint file requires 56 Mb of memory, but if you need to store additional information (on the architecture of the net, training epochs, loss,\dots) the required allocation is around 220 Mb. } stored for the full net occupies 56.14 Mb, whereas that of its reduced versions less than 10 Mb. It can also be noted that the use of a FNN instead of the PCE is saving space in memory of two order of magnitude: $10^{-4}$ against $10^{-2}$.\\
\cref{cifar:res} shows also that for the POD+FNN case the net is not requiring an additional training with the whole dataset since after the initialization, i.e. at epoch 0, its accuracy is acceptable and for index 7 is also already high. The immediate consequence of this is the saving of the time needed to gain a performing network, which is in the order of 5 hours. It can be seen that these considerations are consistent using a different set of data, as the custom dataset in exam. \cref{custom:res} reports also how after the initialization POD+FNN has already a greater accuracy than VGG-16 for all the choices of $l$. \\
For both cases, it can be observed that the proposed reduced CNN achieved a similar accuracy (in most cases also greater) as the original VGG-16 but with much smaller storage. Furthermore, increasing the cut-off layer index $l$ leads to higher accuracy since we are retaining more original features, but on the other hand there is a smaller compression ratio. For this reason, as pointed out before, the right choice for $l$ is a trade-off between the level of accuracy and the reduction, and depends also on the field of application.


%% file: sections/conclusion.tex
In this paper we proposed a generic framework for reduction of neural networks, which aims to reduce the number of layers in the net at the expense of a minimal error in the final prediction. Such a reduction occurs by replacing a finite set of the network layers with a response surface, involving also dimensionality reduction techniques to operate on a low-dimensional space. We analyzed different dimensionality reduction methods, and we investigated how the combination of these techniques with different input-output mappings can lead to differences in the final accuracy.

The creation of this reduced network has one main goal: the compression of existing deep neural network architecture in order to be included into an embedded system with memory and space constraints. The numerical experiments on a convolutional neural network show that the proposed techniques can produce a reduced version (in terms of number of layers and parameters) of an existing network with a saving in memory allocation, keeping the same good level of accuracy of the original CNN. Furthermore, from the results it emerges that the methodology combining POD with FNN leads also to a decrease in training time, which makes the proposed framework better than the inspiring method proposed in~\cite{cui2020active}.

The main drawback of this technique is the necessity to start with an already trained network for reducing it. Indeed, despite the saved space and memory, the learning procedure in many problems results the real bottleneck. In these cases, our framework could be extended in order to reducing the architecture dimension during the training, and not only once it is finished, inducing hopefully a remarkable speedup in the optimization step.